\documentclass[conference]{IEEEtran}
\ifCLASSINFOpdf
\else
\fi

\usepackage{amsmath}
\usepackage{graphicx}
\usepackage[]{algorithm2e}
\usepackage{multirow}
\newcommand{\reals}{{\mbox{\bf R}}}
\newcommand{\sign}{\mathop{\bf sign}}
\newcommand{\relu}{\mathop{\bf ReLU}}
\newcommand{\ones}{\mathbf 1}

\newcommand{\minuseq}{\mathrel{{-}{=}}}
\usepackage[top=18mm, bottom=18mm, left=15mm, right=15mm]{geometry}
\begin{document}
\RestyleAlgo{boxed}
%
\title{SparseNN: An Energy-Efficient Neural Network Accelerator Exploiting Input and Output Sparsity}

\author{\normalsize
  \large Jingyang Zhu$^1$, Jingbo Jiang$^1$, Xizi Chen$^1$ and Chi-Ying Tsui$^2$ \\
  \small
  Department of Electronic and Computer Engineering, Hong Kong University of Science and Technology, Hong Kong\\
  \small
  Email: $^1$\{jzhuak, jingbo.jiang, xchenbn\}@connect.ust.hk, $^2$eetsui@ust.hk
}


%


\maketitle

\begin{abstract}
Contemporary Deep Neural Network (DNN) contains millions of synaptic connections with tens to hundreds of layers.
The large computation and memory requirements pose a challenge to the hardware design. In this work, we
leverage the intrinsic activation sparsity of DNN to substantially reduce the execution cycles and the energy consumption.
An end-to-end training algorithm is proposed to develop a lightweight run-time predictor for the output activation sparsity on the fly.
From our experimental results, the computation overhead of the prediction phase
can be reduced to less than 5\% of the original feedforward phase with negligible accuracy loss.
Furthermore, an energy-efficient hardware architecture, SparseNN, is proposed to exploit both the input and output sparsity.
SparseNN is a scalable architecture with distributed memories and processing elements connected through a dedicated on-chip network.
Compared with the state-of-the-art accelerators which only exploit the input sparsity, SparseNN can
achieve a 10\%-70\% improvement in throughput and a power reduction of around 50\%.
\end{abstract}


%
\IEEEpeerreviewmaketitle

\section{Introduction}
Deep Neural Networks (DNNs) are one of the fundamental machine learning models. In the past decade, DNNs have attracted great
research interest due to their promising results in various domains, including visual recognition \cite{krizhevsky2012imagenet},
natural language processing \cite{amodei2016deep}, and artificial intelligence \cite{silver2016mastering}. Although DNNs can outperform
many traditional machine learning models, the large computation and storage requirements pose an obstacle to the extensive deployment
in embedded applications. For instance, Inception-v4 \cite{szegedy2017inception}, the latest Google DNN model for visual recognition,
requires 12GMACs and more than 42M parameters for classifying a single frame. Therefore, considering the limited resources
in the embedded platform, both algorithmic and architectural optimizations are
required to deliver an energy-efficient solution for DNNs.

In order to avoid overfitting and to be biologically plausible, Rectified Linear Unit (ReLU) is extensively used in DNNs, which
leads to a large amount of zeros in the activations of hidden layers \cite{dahl2013improving}. It is reported
that there is around 50\% sparsity in the contemporary DNN models \cite{albericio2016cnvlutin}. The zero activations can be exploited
for the design of an energy-efficient implementation as the multiplications and memory access associated with these zero activations can be
safely bypassed without affecting the performance.
The activation sparsity can be classified into two categories: the input activation sparsity and
the output activation sparsity. The input activation sparsity refers to the zero activations within the input feature map, and
it is always known when the computation starts. On the other hand, the output activation sparsity,
indicating the zero activations in the output feature map, is unknown until the computation of the current layer is finished.
In this work, we propose an efficient end-to-end training algorithm to form a run-time predictor that can predict the output
activation sparsity before the actual computation of the current layer.
The computation overhead of making the prediction is less than 5\% of the original feedforward calculation.
To efficiently exploit these sparsity to achieve high energy-efficiency, a specialized hardware architecture, SparseNN is proposed.
SparseNN is a scalable Network-on-Chip (NoC) based architecture with distributed processing elements and memories. It can effectively
take the advantage of both input and output activation sparsity.
From the experimental results, it is shown that the throughput
can be improved by 10\%$\scriptsize{\sim}$70\% with a power reduction of 50\% when these two types of sparsity are jointly utilized.

\section{Related Work}
Different optimization techniques have been proposed to improve the energy-efficiency of the deep learning accelerator.
The DianNao series illustrate a series of specialized architectures for the deep learning  \cite{chen2014diannao}, \cite{chen2014dadiannao}.
The customized datapath including multiplier arrays, adder trees, and nonlinear units,
shows a superior performance over the conventional computing platforms like CPU and GPU.
Cnvlutin \cite{albericio2016cnvlutin} enhances the computation scheduling in DianNao by deliberately skipping the zero input
activations. Han et al. \cite{han2015deep} proposed the deep compression algorithm to significantly reduce the memory footprint of DNNs.
A specialized hardware architecture, EIE, was designed to accelerate the inference phase of the compressed models in \cite{han2016eie}.
Davis et al. \cite{davis2013low} adopted singular value decomposition (SVD) as the output sparsity predictor to reduce the
computation complexity in the feedforward pass.
Based on that, an architecture, LRADNN, was proposed to utilize the output sparsity to improve the
energy efficiency by bypassing the zero output activations in \cite{zhu2016lradnn}.
However, none of the previous works leverage both the input and output activation sparsity at the same time.
In summary, this work brings the following contributions:
\begin{itemize}
  \item A novel end-to-end training algorithm is proposed to generate the output sparsity predictor of the neural network.
        The scalability and the predicted sparsity are improved compared with the previous SVD approach.
  \item A scalable NoC based architecture is developed to take advantage of both input and output activation sparsity.
  \item The proposed architecture is implemented in ASIC and simulations are carried out to verify the improvement in throughput and energy
        consumption using three real benchmarks.
\end{itemize}

\section{Preliminaries}
\subsection{Neural Networks}
\begin{figure}
  \centering
  \includegraphics[width=0.85\columnwidth]{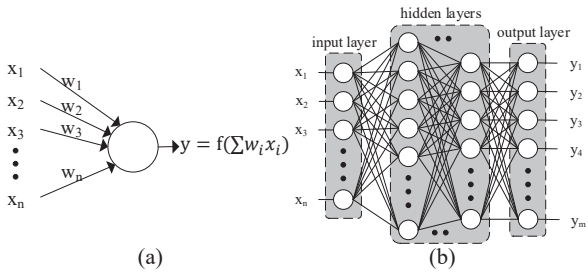}\\
  \caption{(a) The arithmetic computation associated with one neuron. (b) The layer-wise structure of the DNNs.}\label{fig/DNN}
\end{figure}

Neural networks are usually represented as a directed acyclic graph (DAG), where each node refers to a neuron
and the synaptic connection between two neurons is represented by an edge. The neural networks are usually organized
into a stack of hierarchical layers as shown in Fig.~\ref{fig/DNN}.
The layer-wise structure is biology-inspired and is demonstrated to have
algorithmic superiority in real applications. Generally speaking, the first several layers usually act as the low-level feature
extractors (e.g. edges), and the last several layers are able to represent the high-level features (e.g. complex contour).
The arithmetic computation associated with a neuron is the weighted sum of the connected input activations as shown in Fig.~\ref{fig/DNN},
and the layer-wise computation can be compressed into the following concise matrix-vector function:
\begin{equation}\label{eq_ff}
  a^{(l+1)} = f(W^{(l)} a^{(l)})
\end{equation}
where $a^{(l)}$ is the input activation vector, $a^{(l+1)}$ is the output activation vector to be computed, $W^{(l)}$ is the weight
matrix, and $f$ is a nonlinear function (ReLU is typically used). By iteratively applying Eq.~\eqref{eq_ff} following
the topological order of the DAG, we can obtain the final prediction results of the neural network.
Such feedforward pass of DNNs is usually known as the inference phase. On the other hand, the training phase of the DNNs
is often termed as the backpropagation as it is conducted in the reversed order of the feedforward pass.

\subsection{Sparsity Predictor}
\begin{figure}
  \centering
  \includegraphics[width=0.85\columnwidth]{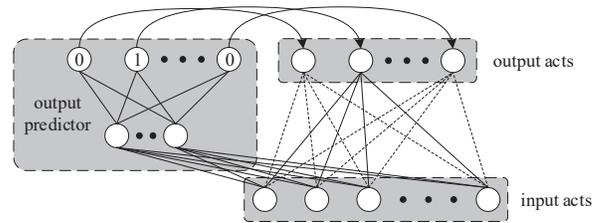}\\
  \caption{The output activation sparsity predictor of DNNs. Only the nonzero output activations are scheduled for the feedforward computation (solid line). The predicted zero output activations are bypassed (dotted line). }\label{fig/predictor}
\end{figure}

Due to the nature and the structure of DNNs, there are a large amount of zeros exist in the input and output activations \cite{albericio2016cnvlutin}.
The input activation sparsity can be easily exploited using
the leading nonzero detector \cite{han2016eie} because $a^{(l)}$ is already known in the feedforward pass when Eq.~\eqref{eq_ff} needs to be computed.
However, the output sparsity is not known until the calculation of the current layer is finished. A common technique to exploit the output sparsity is to add a new
prediction phase, with lightweight computation complexity, to predict whether the output is zero as shown in Fig.~\ref{fig/predictor} \cite{davis2013low}.
Before the feedfoward computation using the original weight matrix is begun, the activeness of each neuron in the output
layer is predicted using a lightweight output predictor.
Based on the predicted output result, we only execute the feedforward computations associated with the
neurons that are predicted as nonzero while the operations of the others are bypassed. Since the prediction phase is a lightweight
procedure and the amount of the bypassed computations is significant, the overall computation complexity is reduced.

In \cite{davis2013low}, the low rank approximation of the weight matrix is used as the output sparsity predictor of the DNNs.
More specifically, the weight matrix $W^{(l)} \in \reals^{m \times n}$ is decomposed into the product of two low rank matrices
$U^{(l)} \in \reals^{m \times r}$ and $V^{(l)} \in \reals^{r \times n}$.
$U^{(l)}$ and $V^{(l)}$ are the first $r$ left-singular vectors and right-singular vectors of $W^{(l)}$, respectively.
The feedforward pass of DNNs using SVD output predictor can then be summarized in the following equations:
\begin{align}
  p^{(l+1)} &= \sign(U^{(l)}V^{(l)}a^{(l)}) \label{eq_ff_out_predictor}\\
  a^{(l+1)} &= p^{(l+1)} \circ f(W^{(l)} a^{(l)}) \label{eq_ff_out_sparse}
\end{align}
where $p^{(l+1)}$ is the sparsity predictor of the output activations and $\circ$ represents the Hadamard product between
the predictor and the original feedforward pass.
The computational complexity of the truncated SVD scheme is $\mathcal{O}(r(m+n))$, which is smaller than the complexity of the
original feedforward computation (i.e. $\mathcal{O}(mn)$) when $r$ is much smaller than $m$ and $n$.
Depending on the polarity of $p^{(l+1)}$, only the positive output neurons are calculated in the subsequent feedforward pass.
However, the truncated SVD scheme is not a good sparsity predictor. More specifically, it always looks for
a solution with the minimum difference of Frobenius norm \cite{davis2013low} but it may not be an optimal sparsity predictor.
For instance, $0.1$ and $-0.1$ are closed to each other in terms of
Frobenius norm, but they give opposite polarity predictions. Moreover, $U^{(l)}$ and $V^{(l)}$ are only updated once-per-epoch in
the training \cite{davis2013low}. The static updating rule limits the flexibility of the backpropagation.

\section{Sparsity Predictor: End-to-End Training}
In order to address the issue of the truncated SVD approach, we propose a more powerful end-to-end training algorithm to search
for a better solution for the output sparsity predictor. It is noted that the internal structure
of the predictor keeps the same as \cite{davis2013low}, i.e. it is based on a pair of $U^{(l)}$ and $V^{(l)}$.
However, the way to come up with $U^{(l)}$ and $V^{(l)}$ is different. Instead of using SVD, they are derived from an end-to-end training phase.

During training we need to backpropagate the gradient of loss $\ell$ into not only the original feedforward pass but also the
sparsity predictor pass.
The gradients are derived iteratively from the output layer to the input layer using the chain rule.
Most of the derivative calculation is straightforward except the passing of the derivative from the predictor to $U$ and $V$.
In Eq.~\eqref{eq_ff_out_predictor}, the derivative of the $\sign$ function will block the output gradient propagate back to
$U$ and $V$ during the backpropagation
since the value of the derivative is zero for all input except when it is 0. Inspired by \cite{courbariaux2016binarized},
we adopt a similar approach using the ``straight-through estimator''. The basic idea is to approximate the $\sign(x)$ with the piece-wise
linear function $\max(-1, \min(1, x))$, whose derivative is $1$ when the input is in $[-1, 1]$. The overall end-to-end training algorithm is
summarized in Alg.~\ref{alg_sparsity_train}. It has three steps: (1) A feedforward step to calculate the activations at each layer;
(2) A backpropagate step to calculate the error term at each layer and the gradients with respect to the parameters;
(3) A gradient descent step to update the trainable parameters.
\begin{algorithm}
\small
\SetKwInOut{Input}{Input}\SetKwInOut{Output}{Output}
 \Input{Network with trainable parameters $\{U^{(l)}, V^{(l)}, W^{(l)}\}_{l=1}^L$}
 \Output{Network with the output sparsity predictor for inference}
 \tcp{Step 1: Feedforward pass}
 \For{$l \leftarrow 1$ \KwTo $L-1$} {
    $a^{(l+1)}_{\mathrm{ori}} = \relu(W^{(l)} a^{(l)})$\;
    $p^{(l+1)} = \sign(U^{(l)}V^{(l)}a^{(l)})$\;
    $a^{(l+1)} = p^{(l+1)} \circ a^{(l+1)}_{\mathrm{ori}}$\;
 }
 \tcp{Step 2: Backpropagate pass}
 Compute $\delta^{(L)} = \frac{\partial \ell}{\partial a^{(L)}}$ knowing $a^{(L)}$ and $a^*$\;
 \For{$l \leftarrow L-1$ \KwTo $1$} {
    $\frac{\partial \ell}{\partial p^{(l+1)}} = \delta^{(l+1)} \circ a^{(l+1)}_{\mathrm{ori}}$\;
    $\frac{\partial \ell}{\partial a^{(l+1)}_{\mathrm{ori}}} = \delta^{(l+1)} \circ p^{(l+1)}$\;
    $\theta^{(l)} = \frac{\partial \ell}{\partial U^{(l)}V^{(l)}a^{(l)}} = \frac{\partial \ell}{\partial p^{(l+1)}}\ones_{|U^{(l)}V^{(l)}a^{(l)}|<1}$\;
    $\gamma^{(l)} = \frac{\partial \ell}{\partial W^{(l)}a^{(l)}} = \frac{\partial \ell}{\partial a^{(l+1)}_{\mathrm{ori}}}\ones_{W^{(l)}a^{(l)}>0}$\;
    $\delta^{(l)}=\frac{\partial \ell}{\partial a^{(l)}}=(W^{(l)})^T \gamma^{(l)}$\;
 }
 \tcp{Step 3: Stochastic gradient descent}
 \For{$l \leftarrow 1$ \KwTo $L-1$} {
    $\frac{\partial \ell}{\partial U^{(l)}} = \theta^{(l)}(V^{(l)}a^{(l)})^T$;
    $\frac{\partial \ell}{\partial V^{(l)}} = (U^{(l)})^T \theta^{(l)}(a^{(l)})^T$\;
    $\frac{\partial \ell}{\partial W^{(l)}} = \gamma^{(l)} (a^{(l)})^T$\;
    $(U^{(l)}, V^{(l)}, W^{(l)}) \minuseq \eta(\frac{\partial \ell}{\partial U^{(l)}}, \frac{\partial \ell}{\partial V^{(l)}}, \frac{\partial \ell}{\partial W^{(l)}})$\;
 }
 \caption{The proposed End-to-End training algorithm for DNNs with the output sparsity predictor.}
 \label{alg_sparsity_train}
\end{algorithm}
To regularize the sparsity of the output activations, we add the $\ell_1$ norm of the sparsity predictor $p^{(l)}$ to the original loss function
to optimize both error rate and sparsity level during training.
Therefore, the gradients with respect to $p^{(l+1)}$ in Alg.~\ref{alg_sparsity_train} will be modified to:
\begin{equation}\label{eq_bp_l1}
  \frac{\partial \ell}{\partial p^{(l+1)}} = \delta^{(l+1)} \circ a^{(l+1)}_{\mathrm{ori}} + \lambda \sign(p^{(l+1)})
\end{equation}
where $\lambda$ is the regularization factor controlling the sparsity of the predictor.

\section{SparseNN: A Scalable Hardware Architecture}
After the output sparsity predictor $U^{(l)}$ and $V^{(l)}$ are obtained using the proposed end-to-end training algorithm,
a specialized hardware architecture is required to accelerate the inference phase of the DNNs with both input and output
sparsity. Traditional Single Instruction Multiple Data (SIMD) microarchitecture like \cite{zhu2016lradnn}\cite{whatmough201714}
is not a scalable solution because the memory bandwidth increases linearly with the SIMD width.
In addition, as for each memory access, several consecutive weights are read out. However, due to the sparsity, not all weights are
used in the computation of outputs. The redundant memory access reduces the energy efficiency and at the same time, some of the processing
elements become idle and this will affect the overall throughput also.
To exploit the input sparsity, a distributed hardware accelerator, called EIE, targeted for
accelerating DNNs with compressed weights was proposed in \cite{han2016eie}.
In this work, we adopt the basic microarchitecture of EIE and enhance it to exploit both the
input and output sparsity. The proposed architecture, SparseNN, has the following distinct features apart from EIE:
(a) A buffered NoC flow control leads to a more efficient use of the on-chip routing fabrics;
(b) A sparsity predictor is added and a different computation schedule is developed for computing the predictor;
(c) Additional skipping blocks are designed for both input and output sparsity.

\subsection{Hierarchical Architecture of SparseNN}
\begin{figure}
  \centering
  \includegraphics[width=0.99\columnwidth]{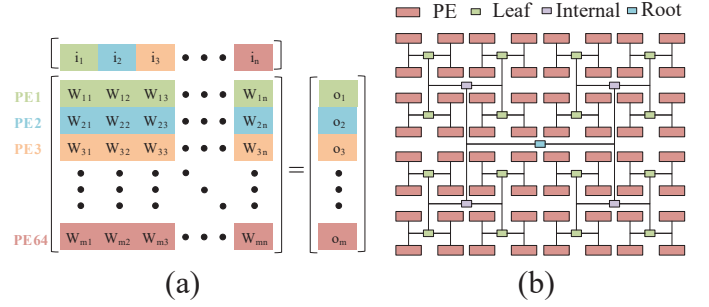}\\
  \caption{(a) The row-based interleaving of weights and activations in SparseNN. (b) The hierarchical structure of SparseNN with 64 processing elements and 3-level routing fabrics. }\label{fig/sparsenn_top}
\end{figure}
SparseNN is a scalable distributed hardware architecture consisting of 64 processing elements (PEs).
As shown in Fig.~\ref{fig/sparsenn_top}, 64 PEs are connected through a dedicated 3-level on-chip H-tree network, which has routers
at the leaf-level, the internal-level, and the root-level.
The computation of the matrix-vector multiplication of Eq.~\eqref{eq_ff} is distributed to each PE.
More specifically, all rows of the weight matrix $W_{(j, :)}$, and the input activations $i_j$ are
stored in the $k^{\mathrm{th}}$ PE, and output activations $o_j$ are computed by the $k^{\mathrm{th}}$ PE, where $j\mod64 = k$.
Since each PE only stores a subset of the
input activations, the output activation can not be computed locally until all the input activations are received. As a result,
an additional broadcasting stage is required to distribute the local input activations stored in each PE
along with the input indices to all PEs through the on-chip network.
In order to exploit the input sparsity, only the nonzero activations in the PE will be broadcasted.
Each PE starts the local multiplication and accumulation of the input activations as soon as it receives the nonzero input activations from
the on-chip network.
During the inference computation, SparseNN is first used to calculate the sparsity predictor (i.e. Eq.~\eqref{eq_ff_out_predictor}) and then
the original matrix-vector multiplication in Eq.~\eqref{eq_ff_out_sparse} is computed. Since the dimensions of the matrices $U$, $V$ and $W$
are very different, different schedulings for computing these matrices are needed and will be discussed in Section V.C.

\subsection{On-chip Network Design}
In EIE, the timing overhead of broadcasting the input activations to the PEs does not cause degradation in performance.
Since the dimension of matrix $W$ is usually very large and multiple rows are mapped to each PE, so whenever the PE
receives an input activation, it will take multiple cycles to compute the multiplications with the weight of each mapped
row and the next input activation will only be needed many cycles later. Therefore, it has enough time margin for the next
broadcasting input activation to arrive to avoid idling cycles.
However, in a general accelerator for DNNs, the weight matrix may not necessarily be a square one.
For example, the $V$ matrix of the sparsity predictor is usually a matrix with fat shape.
Very few output activations are mapped to each PE if the weight matrix has a smaller number of rows, and hence
for each PE, it only takes a few cycles to consume the received input activation.
So if the next input activation does not arrive on time, there will be idling cycles and affect the overall performance.
As a result, the on-chip network of SparseNN is deliberately designed to make sure the activation can arrive
every clock cycle to keep the datapath in the PE always busy. Here we adopt a general buffered flow control of
the on-chip network. Four nonzero input activations are arbitrated at each level of the routing node. The activation with the smallest
index will be granted to the next level while the others will be stored in the buffer at the current node, waiting for the arbitration
in the next cycle. The transmission of activations is fully pipelined, and hence each PE can receive the data every cycle.
However, the arriving input activation is \textbf{out-of-order}, meaning
the index of the received nonzero input activations at each PE may not follow a strictly increasing order.
This is because arbitration is performed locally at each routing level.
The earlier nonzero activations might be blocked in a leaf node, while some of the activations with a higher index may
enter into a higher level node from another leaf node.
However, the out-of-order input activations do not affect the computation results as the matrix-vector multiplication is commutative and the receiving order is not important.
The buffered flow control needs additional temporary storage in the routing node but as shown in Section VI,
the routing logic takes less than 1\% of the total area of SparseNN and this additional overhead is negligible.

\subsection{Computation Schedule for Sparsity Predictor}
\begin{figure}
  \centering
  \includegraphics[width=0.95\columnwidth]{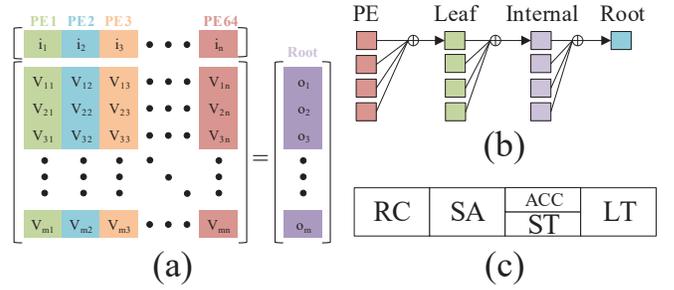}\\
  \caption{(a) The column-based interleaving of $V$ with the partial sum of each row is calculated in 64 PEs. (b) The accumulation of partial sum is conducted through 3-level routing nodes. (c) The modified pipeline stage in NoC router: Routing Computation (RC), Switch Allocation (SA), Switch Traversal (ST), ACCumulation (ACC), and Link Traversal (LT).}\label{fig/columnwise_schedule}
\end{figure}

In the original computation scheduling of EIE, each row of the weight matrix $W$ is distributed to one of the 64 PEs, and
the corresponding output activations are calculated. We call this the row-based scheduling.
The computation of the sparsity predictor $U$ and $V$ can also be conducted in a similar way but then the hardware
utilization will not be optimized.
If the row number of the weight matrix is smaller than 64, not all PEs are mapped with the output activations under the row-based scheduling.
This situation happens for the $V$ matrix in the sparsity predictor because the rank size $r$ is typically smaller
than 64. In order to address this limitation of row-based scheduling of the matrix-vector multiplication, we propose
another column-based scheduling as shown in Fig.~\ref{fig/columnwise_schedule}.
In Fig.~\ref{fig/columnwise_schedule}(a), the columns (instead of rows) of $V$ are interleavedly mapped to the 64 PEs.
Each PE calculates the partial sum of the output activations $o$ on the right hand side. The accumulation of the partial sums is conducted
through the 3-level H-Tree in Fig.~\ref{fig/columnwise_schedule}(b), and the final results of the output activations are stored in
the root node. The accumulation operation is embedded in the 4-stage pipelined router shown in Fig.~\ref{fig/columnwise_schedule}(c).
The utilization rate of the $V$ computation is closed to 100\% even when the rank size $r$ is as low as 16.
The following $U$ computation stage in the sparsity predictor uses the original row-based scheduling as the row number is the same as the
number of output activations of $W$, which is usually much greater than 64.

\subsection{Architecture of PE with the Output Sparsity Bypass}
\begin{figure}
  \centering
  \includegraphics[width=0.95\columnwidth]{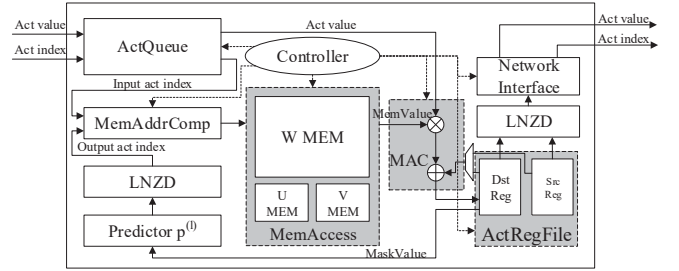}\\
  \caption{The microachitecture of the processing element. The major blocks of the PE include the input activation queue (ActQueue), the leading nonzero detector (LNZD), the memory address computing unit (MemAddrComp), the memory (MemAccess), the multiplier-accumulator (MAC), and the activation register file (ActRegFile). }\label{fig/pe}
\end{figure}

The architecture of the PE in SparseNN is shown in Fig.~\ref{fig/pe}.
The datapath of the PE consists of 5 pipeline stages: memory address computation, memory access, multiplication, addition, and write back.
The two physical register files are organized as a pair of ping-pong buffers, which alternatively act as the source and destination register files from layer to layer.
A complete computation flow of the PE undergoes three matrix-vector computation phases for $V$, $U$, and $W$, respectively.
\subsubsection{$V$ computation phase}
The local nonzero input activation $a_j$ and its associated index $j$ are scanned from the
source register file which stores all local input activations,
and pushed into the datapath. The column-based scheduling in Fig.~\ref{fig/columnwise_schedule}
is then proceeded to calculate the partial sum in each row. When the partial sum of one row is finished, the result will be
sent to the on-chip network for the accumulation. The root node receives the final accumulated result of $V$ computation and
broadcasts it back to all 64 PEs. The results will be temporarily stored in the activation queue if the PE has not finished the
$V$ computation.
\subsubsection{$U$ computation phase}
With the received $V$ results, the row-based scheduling (i.e. Fig.~\ref{fig/sparsenn_top}) of $U$ computation is conducted
in each PE. In each clock cycle, the PE only processes the head of activation queue, and pushes the locally-stored rows of the $U$ matrix
and the results of $V$ computation phase to the datapath.
At the end of the $U$ computation phase, the output sparsity predictor $p^{(l)}$ is stored in a
dedicated 1-bit register bank.
\subsubsection{$W$ computation phase}
The local nonzero input activation $a_j$ and its associated index $j$ are scanned from the source register file, and broadcasted to
all the PEs through the H-Tree. After receiving the nonzero input activation and the index, each PE
then multiplies the received input activation with the weights of all output activations mapped to the PE that are predicted by the sparsity
predictor to be nonzero.
In each cycle, the leading nonzero detector of the predictor register bank searches the next nonzero output activation
for computation and the intermediate results are stored in the destination registers.

\section{Experimental Results}
\subsection{Experimental Setup}
We first compare the performance of the proposed end-to-end training algorithm with that of the conventional truncated SVD scheme on MNIST-BASIC dataset (BASIC)
along with two challenging variants \cite{larochelle2007empirical}.
The variation extends the original hand-written digits with the rotation (ROT) and background superimposition (BG-RAND).
Two different neural network architectures are explored in this work: the 3-layer (with 1 hidden layer) and the
5-layer (with 3 hidden layers). Each hidden layer has 1000 neurons.

To evaluate the hardware performance (i.e. area, power, and latency), we implement SparseNN using
Verilog HDL. The functional simulation of the hardware implementation is verified against with the fixed point simulation in Matlab.
SparseNN is synthesized using the Synopsys Design Compiler with TSMC 65nm LP library
under the worst case PVT. To model the area, power, and access time of the memory, CACTI 6.5 \cite{muralimanohar2009cacti} is used.
We collect the toggling rate from the post-synthesize simulation on the real benchmarks and
use it to estimate the power consumption of SparseNN using Synopsys PrimeTime.

\subsection{Performance of the End-to-End Training Algorithm}
\begin{figure}
  \centering
  \includegraphics[width=0.95\columnwidth]{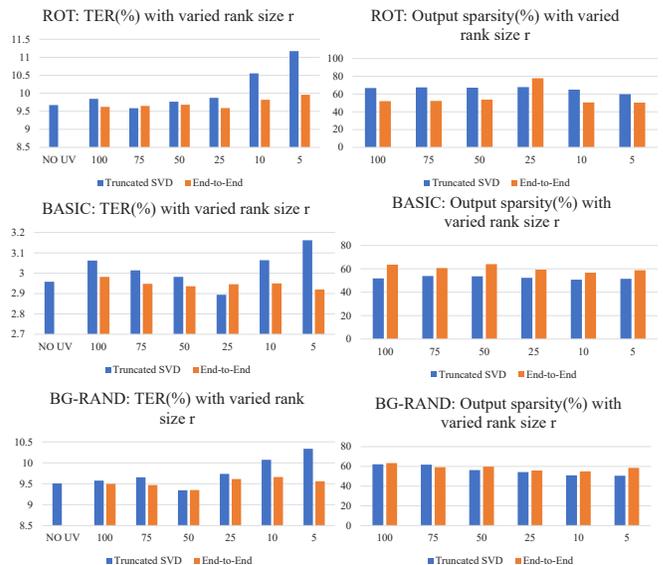}\\
  \caption{Comparison of the proposed end-to-end training algorithm with the truncated SVD scheme on the neural network with one hidden layer. }\label{fig/3_layer}
\end{figure}

The test error rate (TER) and the predicted output sparsity $\rho^{(l)}$ of the 3-layer neural network are shown in Fig.~\ref{fig/3_layer}.
Due to the limitation of space, we only show the results of 5-layer neural network with a rank size 15 in Table.~\ref{Tab/5_layer}, and
the results of the other rank sizes have the similar trend.

\begin{table}[h]
\caption{Test error rate and predicted output sparsity $\rho$ of 5-layer neural network with rank size 15}
\label{Tab/5_layer}
\center
\begin{tabular}{|c|c|c|c|c|c|}
  \hline
  Dataset & Algorithm & TER(\%) & $\rho^{(1)}$ & $\rho^{(2)}$ & $\rho^{(3)}$ \\
  \hline
  \multirow{3}{*}{ROT} & NO UV & \textbf{8.54} & N.A. & N.A. & N.A. \\
  \cline{2-6}
  & SVD & 10.69 & \textbf{90.74} & 28.12 & 34.27 \\
  \cline{2-6}
  & End-to-End & 8.8 & 69.41 & \textbf{64.13} & \textbf{71.07} \\
  \hline
  \multirow{3}{*}{BASIC} & NO UV & 2.738 & N.A. & N.A. & N.A. \\
  \cline{2-6}
  & SVD & 2.728 & \textbf{62.5} & 38.15 & 39.38 \\
  \cline{2-6}
  & End-to-End & \textbf{2.718} & 56.34 & \textbf{65.89} & \textbf{66.7} \\
  \hline
  \multirow{3}{*}{BG-RAND} & NO UV & 10.08 & N.A. & N.A. & N.A. \\
  \cline{2-6}
  & SVD & 10.036 & 51.61 & \textbf{51.49} & 24.01 \\
  \cline{2-6}
  & End-to-End & \textbf{10.03} & \textbf{52.79} & 48.23 & \textbf{41.44} \\
  \hline
\end{tabular}
\end{table}

From Fig.~\ref{fig/3_layer}, we can observe that the proposed end-to-end training algorithm of the sparsity predictor scales
well with the rank size of the $UV$ predictor. For instance, the TER of the truncated SVD scheme is around 1\% larger than the end-to-end
training algorithm in ROT dataset when a small rank size is used. The performance difference is mainly because the $UV$ update is static
in the conventional truncated SVD scheme and cannot be tuned.
The low rank approximation of the weight matrix $W$ is inaccurate when the rank size is small.
In Table.~\ref{Tab/5_layer}, we compare the TER and the output sparsity at each hidden layer of the 5-layer neural
network trained using different algorithms. The network trained by the proposed end-to-end training algorithm
preserves a similar (or even better) accuracy to the SVD approach, but with a higher average sparsity ratio of the hidden layers.
It is mainly because the output sparsity is considered in the end-to-end training algorithm
as we use the $\ell_1$ regularization in the cost function (Eq.~\eqref{eq_bp_l1}).
A higher sparsity ratio is preferred for the better energy-efficiency of SparseNN, because more computation can be skipped.
It is noted that a larger regularization factor $\lambda$ can result in a larger sparsity prediction in each layer, but
TER might be affected due to the underfitting.

\subsection{Performance of the SparseNN}
The design parameters of the mircroarchitecture of the proposed architecture, SparseNN, are listed in Table.~\ref{Tab/microarchitecture}.
\begin{table}[h]
\caption{The mircroarchitecture parameters of 64-PE SparseNN}
\label{Tab/microarchitecture}
\center
\begin{tabular}{c|c}
  \hline
  Micro-architectural parameters & Value \\
  \hline
  Quantization scheme & 16-bit fixed point \\

  On-chip $W$/$U$/$V$ memory per PE & 128KB/8KB/8KB \\

  Activation register no. per PE & 64 \\


  Flow control of NoC router & Packet-buffer with credit \\
  \hline
\end{tabular}
\end{table}
The microarchitecture of SparseNN is inspired by EIE. For instance, the total on-chip weight memory is $128\mathrm{KB}\times64=8\mathrm{MB}$,
and the maximum number of activations in each layer is $64\times64=4\mathrm{K}$. The target critical path of SparseNN is set to 2ns because
the access time of the $128\mathrm{KB}$ SRAM is more than 1.7ns.

The area breakdown of SparsNN is listed in Table.~\ref{Tab/area_breakdown}.
\begin{table}[h]
\caption{The area breakdown of SparseNN by component and by module}
\label{Tab/area_breakdown}
\center
\begin{tabular}{c|c|c}
\hline
& Area ($\mu m^2$) & (\%) \\
\hline
Total & 78,443,365 & (100\%) \\
\hline
Combinational & 1,716,373 & (2.4\%) \\
Buf/Inv & 199,038 & (0.2\%) \\
Non-combinational & 2,068,996 & (2.6\%) \\
Macro (Memory) & 74,426,310 & (94.8\%) \\
\hline
Processing element & 1,216,457$\times$64 & (99.2\%) \\
Routing logics  & 590,062 & (0.8\%) \\
\hline
\end{tabular}
\end{table}
The routing nodes occupy only a small fraction (less than 1\%) of the total area, and the major area contributors are the PEs.
The main reason is the large on-chip SRAMs for $W$, $U$, and $V$ in each PE, which take around $95\%$ of the overall area.

\begin{figure}
  \centering
  \includegraphics[width=0.73\columnwidth]{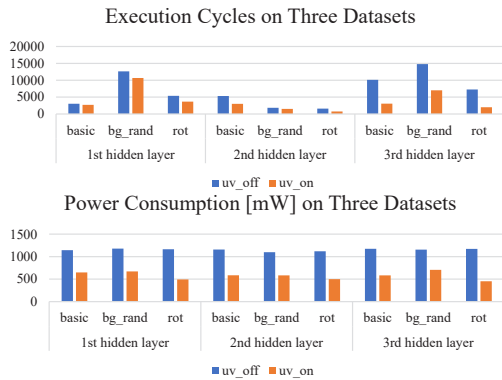}\\
  \caption{Comparison of execution cycles and power consumption on three datasets using the 5-layer DNN. The results are organized in layer-wise, where \emph{uv\_on} and \emph{uv\_off} represent the output sparsity predictor of SparseNN is enabled and disabled, respectively.}
  \label{fig/hardware_results}
\end{figure}

The results on the execution cycle and the power consumption of SparseNN on the three benchmarks are shown
in Fig.~\ref{fig/hardware_results}. When UV predictor is not used, SparseNN is the same as the conventional EIE architecture
which only exploits the input activation sparsity.
From Fig.~\ref{fig/hardware_results}, it can be seen that the improvement in the number of execution cycles
with the output sparsity varies from layer to layer. For the 1st hidden layer, the reduction of cycles
ranges from 10\%$\scriptsize{\sim}$31\%. The inputs to the 1st hidden layer are the same for the UV enabled and the UV disabled networks,
and hence the improvement of throughput only comes from the output sparsity.
The difference of the throughput improvement at different layers and benchmarks is due to the difference in predicted output sparsity.
In addition, the number of nonzero output activations predicted by the sparsity predictor also varies from PE to PE.
For the remaining hidden layers, the reduction of cycles can be as high as 70\%. The predicted output sparsity of the
previous layer will increase the input sparsity of the current layer.
Therefore, the throughput is jointly improved by the input sparsity as  well as the output sparsity.
The improvement in power consumption with output sparsity is almost uniform among all datasets and all hidden layers: around 50\%.
The reasons for the power reduction are twofold: the number of access to the large $W$ memory decreases with the output sparsity,
and the access energy to the $U$, $V$ memory during sparsity prediction phase is small.
We also compare SparseNN with the existing SIMD hardware platforms for DNNs in Table.~\ref{Tab/compare_existing_work}.
\begin{table}[h]
\caption{Comparison with existing SIMD hardware platforms for DNNs}
\label{Tab/compare_existing_work}
\center
\begin{tabular}{|c|c|c|c|}
\hline
Platform & LRADNN \cite{zhu2016lradnn} & DNN-Engine \cite{whatmough201714} & This work \\
\hline
Technology & 65nm & 28nm & 65nm \\
\hline
Peak Perf. & 7.08GOPs & 19GOPs & 64GOPs \\
\hline
$W$ memory & 3.5MB & 1MB & 8MB \\
\hline
Power (mW) & 439$\scriptsize{\sim}$487 & 63.5 & 452$\scriptsize{\sim}$705 \\
\hline
Area ($\mathrm{mm}^2$) & $51$ & $5.76$ & $78$ \\
\hline
\end{tabular}
\end{table}
In SIMD architecture, there is a tradeoff between the parallelism (i.e. SIMD window) and the on-chip bandwidth.
More specifically, the working frequency of LRADNN is slower as the unified memory needs to provide 32 data in each cycle.
On the other hand, the parallelism level in DNN-Engine is limited to 8 in order to achieve a frequency as high as 1.2GHz.
Ideally, DNN-Engine takes $\frac{785\times1000}{8}$ cycles to finish the 1st hidden layer computation
of the dataset BG-RAND. Therefore, the corresponding energy consumption by DNN-Engine is approximately $5.1\mu J$.
On the other hand, the energy consumption of SparseNN for the 1st hidden layer in BG-RAND is around $14\mu J$.
However, as they are implemented in different technology node, to have a fair comparison, we need to scale
the energy consumption accordingly.
From the CACTI memory model, the energy consumption per read access is roughly 11x when the technology node is scaled from
$28\mathrm{nm}$ to $65\mathrm{nm}$ and the memory size changes from 1MB to 8MB.
Therefore, if we take this scaling into account, SparseNN has a 4x better energy-efficiency over the conventional SIMD architecture.

\section{Conclusion}
In this work, we first propose an end-to-end training algorithm to obtain the $U$ and $V$ matrices
for the sparsity predictor from the backpropagation.
The scalability with rank and the predicted sparsity are better than the traditional truncated SVD scheme. Then, a specialized
architecture, SparseNN, is developed to exploit both the input and output sparsity. Our evaluations demonstrate
that with the output sparsity, the throughput of SparseNN can be improved by 10\% to 70\% while the power consumption
is approximately reduced by half. Moreover, SparseNN shows a better scalability and a higher energy-efficiency compared
with the state-of-the-art SIMD architecture.


%
%



%
%
%

\bibliographystyle{unsrt}
{\scriptsize\bibliography{reference}}

\end{document}